\documentclass[letterpaper, 10 pt, conference]{ieeeconf}  

\IEEEoverridecommandlockouts                              

\usepackage{amsmath} 
\usepackage[nolist,nohyperlinks]{acronym}
\usepackage{todonotes}
\usepackage{graphicx}
\usepackage{subcaption}
\usepackage{amsfonts} 
\usepackage{float}
\usepackage{comment}
\usepackage{siunitx}
\usepackage{hyperref}
\usepackage{adjustbox}
\usepackage{caption}
\usepackage{array}
\usepackage{booktabs} 
\usepackage{fontawesome5}
\usepackage{wrapfig}
\usepackage{lipsum}
\usepackage{xspace}

\usepackage{multirow}    
\usepackage{rotating}    

\usepackage[ruled]{algorithm2e}
\SetKwComment{Comment}{/* }{ */}

\acrodef{AI}{Artificial Intelligence}
\acrodef{CAD}{Computer-Aided Design}
\acrodef{NeRF}{Neural Radiance Field}
\acrodef{NGP}{Neural Graphics Primitives}
\acrodef{DoF}{Degrees of Freedom}
\acrodef{SDF}{Signed Distance Field}
\acrodef{MLP}{Multi-Layer Perceptron}

\def\eg{\emph{e.g.,}\xspace} 

\def\ie{\emph{i.e.,}\xspace}

\definecolor{lightgreen}{HTML}{CEEAD6}
\definecolor{lightred}{HTML}{FAD2CF}
\definecolor{lightorange}{HTML}{FEEFC3}
\definecolor{lightblue}{HTML}{30CEFE}
\definecolor{darkerblue}{HTML}{5CA3FF}
\definecolor{brightpurple}{HTML}{9865FE}

\title{\LARGE \bf
The Temporal Trap: Entanglement in Pre-Trained Visual Representations for Visuomotor Policy Learning
} 

\author{\ Nikolaos Tsagkas$^{1,\alpha}$,\ Andreas Sochopoulos$^1$\ Duolikun Danier$^1$,\\ \ Chris Xiaoxuan Lu$^{2,\varepsilon}$, \ Oisin Mac Aodha$^{1,\varepsilon}$\\ 
  $^1$University of Edinburgh, \ $^2$UCL\\
\thanks{*This work was supported by the United Kingdom Research and Innovation (grant EP/S023208/1), EPSRC Centre for Doctoral Training in Robotics and Autonomous Systems (RAS) and ELIAI (Edinburgh Laboratory for Integrated Artificial Intelligence) - EPSRC (EP/W002876/1).\newline
\indent$^\varepsilon$ Indicates equal senior authorship.\newline
\indent$^\alpha$ Corresponding author: N. Tsagkas -- n.tsagkas@ed.ac.uk}
}

\begin{document}

\maketitle
\thispagestyle{empty}
\pagestyle{empty}

\begin{abstract}
The integration of pre-trained visual representations (PVRs) has significantly advanced visuomotor policy learning. 
However, effectively leveraging these models remains a challenge. 
We identify temporal entanglement as a critical, inherent issue when using these time-invariant models in sequential decision-making tasks. 
This entanglement arises because PVRs, optimised for static image understanding, struggle to represent the temporal dependencies crucial for visuomotor control.
In this work, we quantify the impact of temporal entanglement, demonstrating a strong correlation between a policy's success rate and the ability of its latent space to capture task-progression cues. 
Based on these insights, we propose a simple, yet effective disentanglement baseline designed to mitigate temporal entanglement. 
Our empirical results show that traditional methods aimed at enriching features with temporal components are insufficient on their own, highlighting the necessity of explicitly addressing temporal disentanglement for robust visuomotor policy learning.
\end{abstract}

\section{INTRODUCTION}
The integration of pre-trained visual representations (PVRs) into visuomotor robot learning has emerged as a promising alternative to training visual
encoders from scratch.
Despite the promising results of these models in downstream robotic applications, including affordance-based manipulation~\cite{li2024affgrasp}, semantically precise tasks~\cite{tsagkas2024click}, and language-guided approaches~\cite{tsagkas2023vlfields,shen2023F3RM}, their deployment in policy learning is still nascent.

Overall, prior works have identified multiple issues revolving around the deployment of PVRs in robot learning. 
First, a consensus exists that we have not identified a single PVR that consistently leads to the best performance.
This issue concerns not just the task at hand~\cite{parisi2022unsurprising} but also the policy training paradigm~\cite{hu2023pretrainedvisionmodelsmotor}, with performance varying greatly.
Similarly, policy robustness might suffer when the scene undergoes visual perturbations, even though the key motivation behind using PVRs is their generalisation capabilities~\cite{10611331, burns2024what,tsagkas2025eai}.
We add to this list of issues that hinder policy performance by identifying the problem of temporal entanglement. 

In this work, we investigate a paradoxical phenomenon: models celebrated for their robustness on traditional computer vision benchmarks (e.g., classification) tend to underperform in robot learning tasks precisely because of this robustness. 
We show that the representations learned by such models exhibit temporal entanglement, not only between adjacent timesteps, but also over long-range temporal dependencies.
Both entanglements correlate strongly with a model's inability to predict task progression, a metric we find to be a strong predictor of policy success.
Building on this insight, we introduce a simple yet powerful baseline for evaluating temporal disentanglement. 
Using this baseline, we demonstrate that conventional approaches, including feature augmentation, architectural changes to the policy, and alternative pre-training objectives, consistently fall short in addressing the temporal entanglement challenge.

In summary, we make the following contributions:\\
\textbf{1.~Identifying temporal entanglement}. We show that PVRs fall short in encoding temporal cues, even ones trained with temporal objective functions, which has a detrimental effect in visuomotor policy learning.
We further demonstrate that these latent spaces lack a reliable notion of task progression. \\
\textbf{2.~Providing tools to quantify entanglement}. We study both short and long-range temporal entanglement and reveal strong correlations with policy success.
To this end, we propose a probing method that captures both types of entanglement and serves as a strong predictor of downstream performance.\\
\textbf{3.~Proposing a disentanglement baseline}. We identify task-progression perception as a critical missing factor in current PVR-latent spaces and propose a simple yet effective approach to augment them with such a signal, yielding a significant boost in policy performance.  

\begin{figure*}[!t]
\begin{center}
\begin{tabular}{c}
\includegraphics[width=0.98\linewidth]{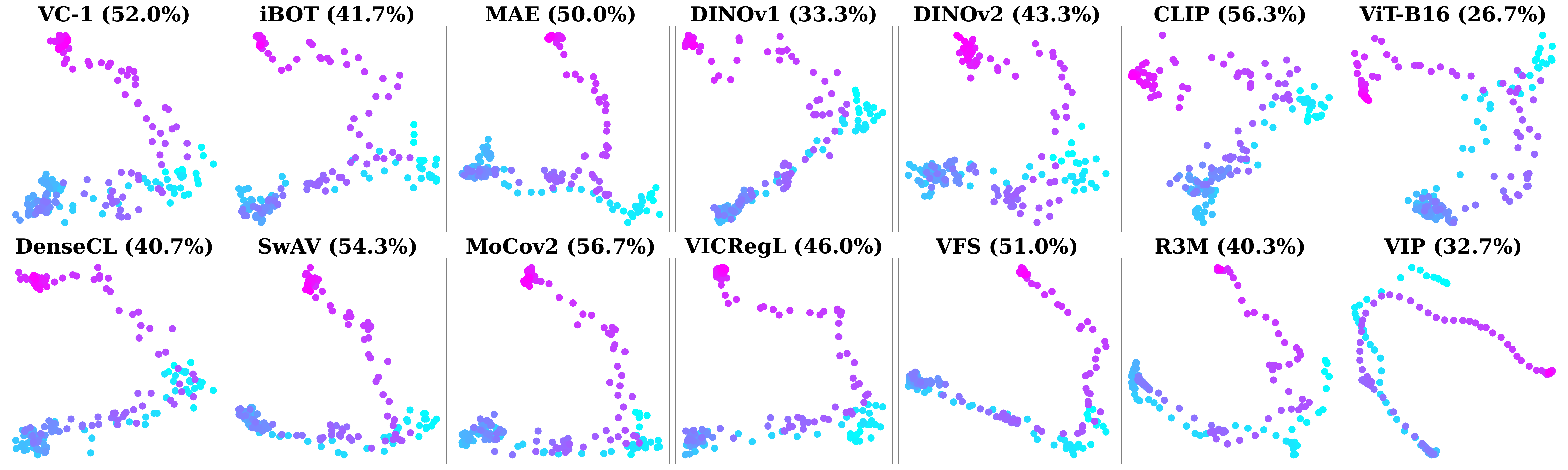} \\
\includegraphics[width=0.98\linewidth]{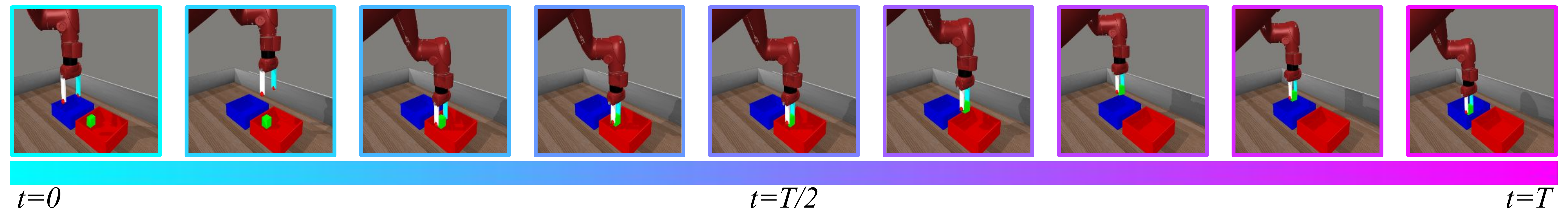} \\
\end{tabular}
\caption{PCA of features from an expert demonstration in Bin Picking across PVRs (Top row: ViT models; Bottom row: ResNet models). Frame colours align with trajectory stages, suggesting feature entanglement during the gripper \textcolor{lightblue}{\textbf{descent}} and \textcolor{brightpurple}{\textbf{ascent}}, and during the \textcolor{darkerblue}{\textbf{gripper stop}} phase. Next to each PVR name we provide the success rate of the corresponding policy for the given task. 
}
\label{fig:pvr_trajectories}
\end{center}
\vspace{-15pt}
\end{figure*}

\section{RELATED WORK}
\label{sec:related_work}
In PVR-based visuomotor policy learning, the incorporation of temporal information remains underexplored. 
Augmentation with temporal perception can happen either at feature level or during training time. 

\textbf{Feature Augmentation}. 
While early fusion methods, such as stacking multiple frames before encoding~\cite{Karpathy_2014_CVPR}, are common in training visual encoders from scratch, late fusion (\ie processing frames individually and stacking their representations~\cite{NIPS2014_00ec53c4}) has shown superior performance with fewer encoder parameters. 
Recent work~\cite{NEURIPS2021_ba3c5fe1} highlights that naive feature concatenation in latent space is insufficient; instead, approaches like FLARE~\cite{NEURIPS2021_ba3c5fe1} incorporate sequential embeddings and their differences, inspired by optical flow techniques. 
Nevertheless, concatenating sequential embeddings as input to policy networks has become standard in visuomotor policy learning~\cite{parisi2022unsurprising} and SoTA generative policies~\cite{chi2023diffusionpolicy,sochopoulos2025fastflowbasedvisuomotorpolicies}. 
However, a gap remains in leveraging PVR features, which are primarily designed for vision tasks, within this temporal framework.

\textbf{Loss Function Augmentation}. 
A major limitation of many PVRs, in the context of visuomotor policy learning, is their inherit lack of temporal perception, as most are pre-trained on static 2D image datasets. 
Temporal perception can be added by employing loss functions that enforce temporal consistency during training (\eg R3M~\cite{nair2022rm} and VIP~\cite{ma2022vip}), when training with video data.  
However, there is no clear consensus on the superiority of this approach compared to alternatives like masked-image modelling (MIM) (\eg MVP~\cite{Xiao2022, hu2023pretrainedvisionmodelsmotor} and VC-1~\cite{NEURIPS2023_022ca1be}). 
This disparity suggests that existing temporal modelling strategies may be insufficient in isolation. 
In later experiments, we evaluate PVRs trained with temporal information and demonstrate that methods trained with a time-agnostic paradigm achieve comparable performance.

We hypothesise that this limitation arises from a lack of task-progression perception, which we address by incorporating positional encoding, a fundamental mechanism in many machine learning approaches.
This straightforward operation has been instrumental in the success of Transformers~\cite{NIPS2017_3f5ee243}, implicit spatial representations~\cite{mildenhall2020nerf}, and diffusion processes~\cite{ho2020denoising}.
Our novelty lies in that we \textit{do not} utilise this tool for encoding the position of an observation in the input stream, but rather in the global temporal stream of the rollout.

\begin{figure}
    \centering

    \includegraphics[width=0.95\linewidth]{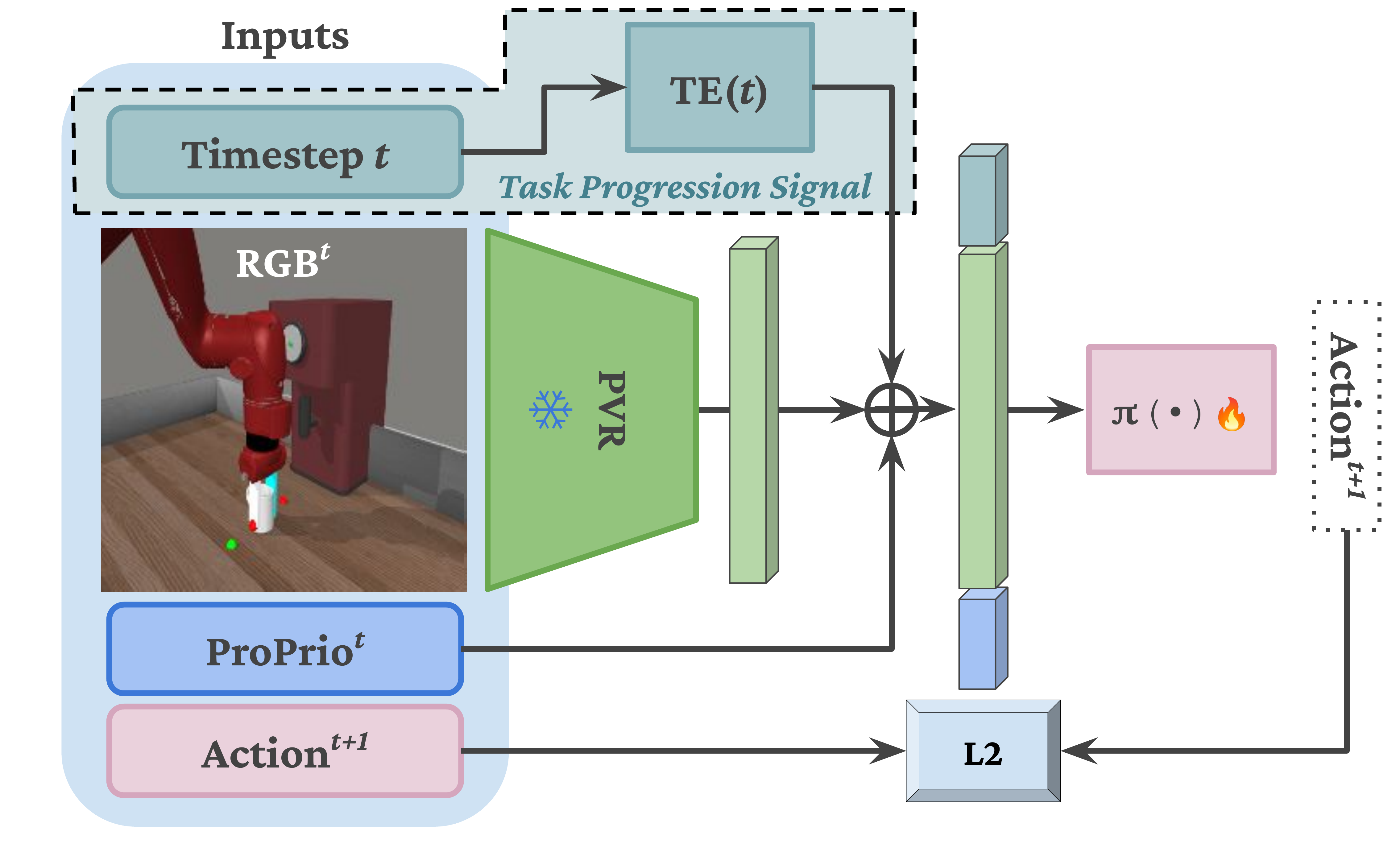}
    \caption{Standard PVR-based behaviour cloning architecture, modified with out task-progression signal. We propose our disentangling baseline, which incorporates a task-progression signal, effectively disentangling features before the policy head.}
    \label{fig:policy}
    \vspace{-15pt}
\end{figure}

\begin{table}[t]
\centering
\caption{Overview of utilised PVRs. \faRobot~indicates models specifically trained for robotic tasks, while~\faStopwatch indicates a temporal training component. Dataset sizes are given in number of images, and  $^\dagger$~denotes number of frames from videos and  $^*$~denotes number of videos. Abbreviations: \textbf{IN}: ImageNet\cite{ridnik2021imagenet21kpretrainingmasses}, \textbf{LVD}: LVD-142M~\cite{oquab2023dinov2}, \textbf{K}: Kinetics~\cite{kay2017kineticshumanactionvideo}, \textbf{E4D}: Ego4D~\cite{grauman2022ego4dworld3000hours}, \textbf{E4D+MNI}~\cite{NEURIPS2023_022ca1be}.  }
\begin{adjustbox}{max width=\linewidth}
\setlength{\tabcolsep}{3pt} 
\footnotesize 
\renewcommand{\arraystretch}{1.1} 
\begin{tabular}{@{}c l l p{2.3cm} l@{}}
\toprule
\textbf{Arch.} & \textbf{PVR} & \textbf{Training Objective} & \textbf{Dataset (Size)} &  \\
\midrule
\multirow{7}{*}{\rotatebox{90}{ViT-B}} 
 & MAE~\cite{He2021MaskedAA}     & MIM              & IN (1.2M)  & \\
 & VC-1~\cite{NEURIPS2023_022ca1be} \faRobot & MIM              & E4D+MNI (5.6M$^\dagger$) & \\
 & DINOv1~\cite{caron2021emerging} & Self-Distillation     & IN (1.2M)                   & \\
 & iBOT~\cite{zhou2021ibot}     & MIM+Self-Distillation & IN (14M)                   & \\
 & DINOv2~\cite{oquab2023dinov2} & MIM+Self-Distillation & LVD (142M)                 & \\
 & ViT~\cite{dosovitskiy2021an}  & Supervised       & IN (14M)                  & \\
 & CLIP~\cite{radford2021learning}  & V-L Contrastive  & LAION (2B)              & \\
\midrule
\multirow{7}{*}{\rotatebox{90}{ResNet-50}}
 & MoCov2~\cite{chen2020mocov2}  & Contrastive      & IN (1.2M)                   & \\
 & DenseCL~\cite{wang2021dense} & Local Contrastive & IN (1.2M)                   & \\
 & SwAV~\cite{caron2020unsupervised} & Clustering       & IN (1.2M)                   & \\
 & VICRegL~\cite{bardes2022vicregl} & VICReg (global+local)   & IN (1.2M)                    & \\
 & VFS~\cite{xu2021rethinking}~\faStopwatch    & Self-Distillation (video) & K (240K$^*$)            & \\
 & VIP~\cite{ma2022vip} \faRobot,\faStopwatch   & Value Function    & E4D (5M$^\dagger$)       & \\
 & R3M~\cite{nair2022rm} \faRobot,\faStopwatch  & Time Contrastive+Language     & E4D (5M$^\dagger$)       & \\
\bottomrule
\label{tab:pvrs_info}

\end{tabular}
\end{adjustbox}
\vspace{-1.29em}
\end{table}

\section{PRELIMINARIES}
\subsection{Imitation Learning via Behaviour Cloning}
\label{ssec:method_preliminaries}

We consider an expert policy \(\pi^\star : \mathcal{P} \times \mathcal{O} \to \mathcal{A}\), which maps a robot's proprioceptive observation \(p \in \mathcal{P}\) and visual observation \(o \in \mathcal{O}\) to an action \(a \in \mathcal{A}\). This policy generates a dataset \(\mathcal{T}^\text{e} = \{(p_t^i, o_t^i, a_t^i)_{t=0}^{T}\}_{i=1}^N\) of \(N\) expert trajectories, where each trajectory  contains \(T\) steps of observations and actions for a task.

We employ behaviour cloning to learn a policy \(\pi_\theta\), parameterised by \(\theta\), to imitate \(\pi^\star\) by minimizing the action discrepancy over demonstrations: 

\begin{equation}
    \mathbb{E}_{(p_t^i, o_t^i, a_t^i) \sim \mathcal{T}^e} \|a_t^i - \pi_\theta(f_\text{PVR}(o_t^i), p_t^i)\|_2^2,
\end{equation}

\noindent where \(f_\text{PVR}\) is a pre-trained visual representation (PVR) that extracts features from \(o_t^i\).
In visuomotor policy learning, it is common to assume the Markov property, whereby the current observation \(x_t = (p_t, o_t)\) suffices for predicting the next state: \(P(x_{t+1} | x_t) = P(x_{t+1} | x_t, x_{t-1}, \dots, x_0)\). 
This allows tasks to be modelled as Markov decision processes, where each action depends only on the current state, enabling the use of behaviour cloning  under this formulation.

As is common practice in similar work~\cite{parisi2022unsurprising, nair2022rm, hu2023pretrainedvisionmodelsmotor}, we use a shallow (4-layer) MLP with ReLU activations and $\tanh$ before the output to predict the mean $\mu$ of a Gaussian with fixed standard deviation $\sigma$, modelling $\pi_\theta(a\mid o) = \mathcal{N^T}(\mu, \sigma^2)$, where $\mathcal{N^T}$ denotes a truncated Gaussian with support in $[-1, 1]^k$ (see Fig.~\ref{fig:policy}).

\subsection{Implementation Details}
\noindent\textbf{Environment}. We conducted our experiments in simulations built on the MuJoCo~\cite{6386109} physics engine. 
Our core experiments were run on the widely used MetaWorld~\cite{yu2020meta} environment, from which we selected ten tasks, based on their level of difficulty, as identified in prior work on PVR-based visuomotor control~\cite{mete2024questselfsupervisedskillabstractions, hu2023pretrainedvisionmodelsmotor}, as well as from our empirical results.
Using the provided heuristic policy, we generated for each task 25 expert demonstrations, with a maximum of 175 steps per rollout.
Also, for the purpose of validating the generality of our proposed baseline, we verified its success on three robot arm tasks from the OpenAI Robotics Suite~\cite{brockman2016openaigym,plappert2018multigoalreinforcementlearningchallenging}.

\noindent\textbf{PVRs}. 
We conducted extensive evaluations across 14 PVRs, including the most popular vision encoders in the field of visuomotor policy learning, that have led to state-of-the-art performance, as summarised in Tab.~\ref{tab:pvrs_info}.
We included PVRs from two main architecture families (Residual Networks (ResNets)~\cite{he2016deep} and Vision Transformers (ViT)~\cite{dosovitskiy2021an}), maintaining a consistent backbone architecture for each group (ResNet-50 or ViT-B/16, with the exception of DINOv2~\cite{oquab2023dinov2}, which employs a smaller patch size of 14).
In general, we selected PVRs with diverse characteristics, concerning the objective function, training dataset and balance between local and global perception. 
We also aimed to include PVRs that have a strong temporal component (\eg R3M) or simply trained with video data (\ie VC-1), to see if they deal better with temporal entanglement.
Finally, we also briefly study whether VideoPVRs (\ie PVRs meant for video perception, and not singular image processing) could be a solution to the entanglement problem.
For this, we selected three popular, state-of-the-art VideoPVRs: TimeSformer~\cite{timesformer}, ViViT~\cite{ViViT} and VideoMAE~\cite{tong2022videomae}.

\noindent\textbf{Policy training}. 
We deploy a similar behaviour cloning training scheme to~\cite{hu2023pretrainedvisionmodelsmotor}, where we train each policy \textit{(PVR, task)} pair 5 times, without updating the PVR weights, and report the interquartile mean (IQM) success rate.
We train for 80K steps, using mini-batches of 128 samples and the Adam optimizer and learning rate equal to $10^{-4}$. 

\begin{figure*}[th]
\centering
\includegraphics[width=\linewidth]{figures/tasks/tasks_v5.pdf}
\includegraphics[width=\linewidth]{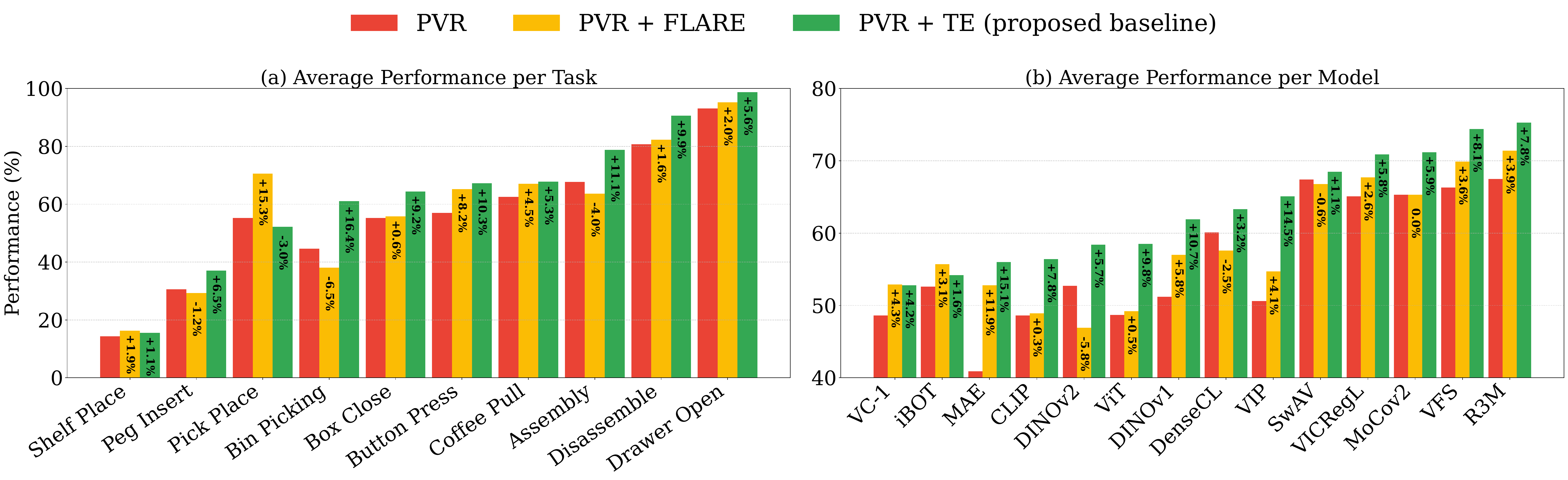}
\caption{Comparison of our Temporal Encoding (TE) against FLARE~\cite{NEURIPS2021_ba3c5fe1} and using no temporal augmentation on PVR features. Results (sorted by TE) show (a) per-task performance and (b) per-model performance. FLARE and TE bars indicate gains over no temporal information.}
\label{fig:results}
\vspace{-0.99em}  
\end{figure*}
\section{TEMPORAL ENTANGLEMENT}
\label{sec:temporal_entanglement}
\subsection{Intuition}
\label{ssec:intuition}
We observe that the assumption of Markovian decision-making in policies using features from frozen PVRs is often invalid. 
This arises because, at each timestep, the available information may be insufficient for the policy to confidently map the current observation to the appropriate action.

Consider the example presented in Fig.~\ref{fig:pvr_trajectories}, where PVR-features of the same pick-and-place trajectory are projected with PCA into 2D. 
Regardless of the PVR utilised, the extracted features seem to suffer from temporal entanglement. 
First, features extracted from the frames where the robot has stopped to pick up the box often form a tight cluster, since the only change is the movement of the gripper fingers, which corresponds to a very small percentage of pixels. 
Second, as the gripper moves down and subsequently ascends, the primary visual change is the cube's vertical displacement relative to the table. 
Consequently, the visual features extracted from the descent and ascent frames may differ only marginally, and only in dimensions affected by the small pixel region of the cube.

Training a policy network to map $(p_t, o_t)$ to $a_t$ becomes difficult under these conditions. 
When multiple observations are nearly indistinguishable, the mapping violates the functional requirement that each input must map to exactly one output. 

\subsection{Short-range Temporal Entanglement}
\label{ssec:short_range_entanglement}
To quantify the short-range temporal entanglement of features extracted from PVRs, we define the \textit{average sequential cosine similarity} across $N_d$ expert demonstrations and $N_T$ tasks. 
Each demonstration trajectory consists of $N$ frames, and each frame is mapped to a feature token $\mathbf{f}_n^{(i,t)} \in \mathbb{R}^d$ for the $n$-th frame of the $i$-th demonstration of task $t$. 
We compute the average feature token for each task:

\[
\bar{\mathbf{f}}^{(t)} = \frac{1}{N_d N} \sum_{i=1}^{N_d} \sum_{n=1}^{N} \mathbf{f}_n^{(i,t)}
\]

\noindent and subtract it to reduce the influence of static visual cues. 
The \textit{average sequential cosine similarity} is then defined as:

\[
\frac{1}{N_d N_T (N-1)} \sum_{t=1}^{N_T} \sum_{i=1}^{N_d} \sum_{n=1}^{N-1} \cos\left( \tilde{\mathbf{f}}_n^{(i,t)}, \tilde{\mathbf{f}}_{n+1}^{(i,t)} \right)
\]

\noindent where $\tilde{\mathbf{f}}_n^{(i,t)} = \mathbf{f}_n^{(i,t)} - \bar{\mathbf{f}}^{(t)}$ is the task-centred feature. 

The left plot of Fig.~\ref{fig:correlations_entanglement} reveals a clear trend: within particular PVR subgroups, such as ResNets, short-range temporal entanglement is strongly and negatively correlated with the policy success rate. This finding substantiates our claim that PVRs producing more diverse sequential tokens during a rollout tend to achieve higher policy success rates.

\subsection{Long-range Temporal Entanglement}
\label{ssec:long_range_entanglement}
We hypothesised that entanglement concerns not only sequential tokens, due to small disparities in the pixel domain, but also from the robot traversing similar areas during the task (\eg ascent/descent during pick-and-place).
To measure the effect of long-range entanglement, we introduce another metric, the \textit{average global cosine similarity}, where for each PVR we measure the average similarity of each token with all others in a rollout, for all expert demos and tasks.  

\[
\frac{1}{N_d N_T N} \sum_{t=1}^{N_T} \sum_{i=1}^{N_d} \sum_{\substack{n,m=1\\n \ne m}}^{N} \cos\left( \mathbf{f}_n^{(i,t)}, \mathbf{f}_m^{(i,t)} \right)
\]

\begin{figure*}[t] 
    \centering
    \begin{subfigure}[b]{0.495\textwidth} 
        \centering
        \includegraphics[width=\linewidth]{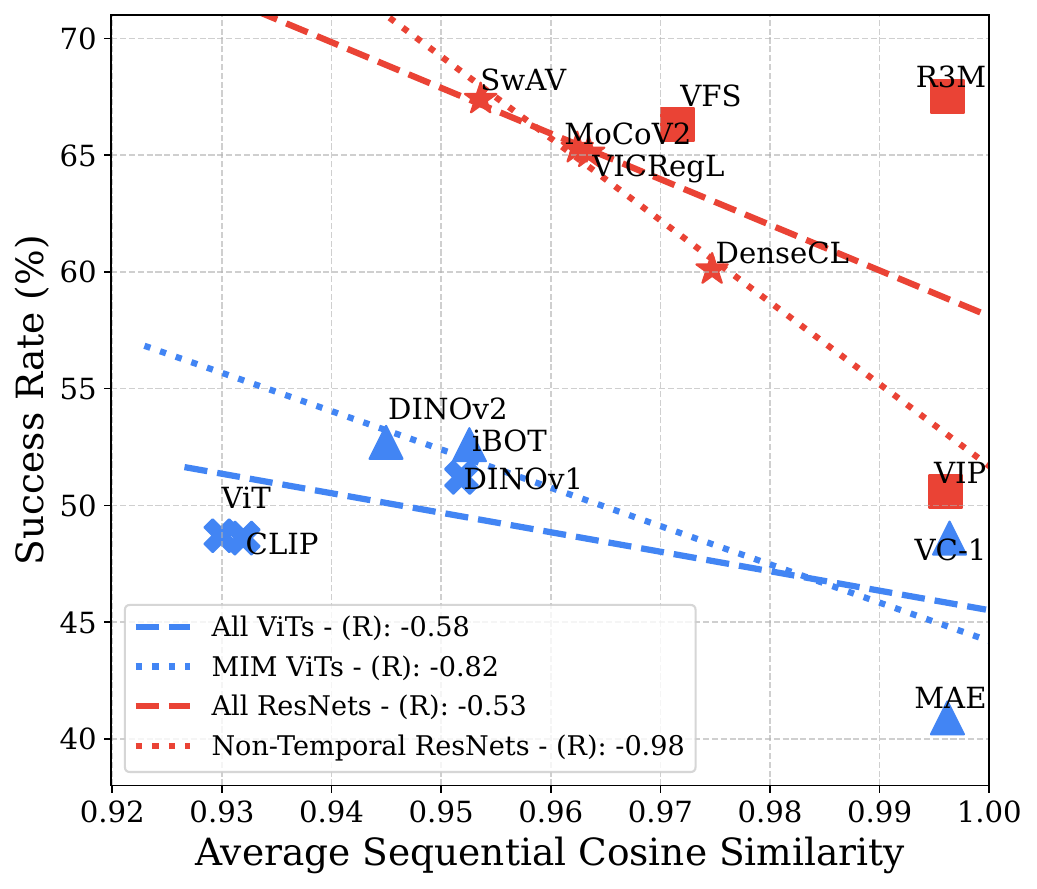}
    \end{subfigure}
    \hfill 
    \begin{subfigure}[b]{0.495\textwidth} 
        \centering
        \includegraphics[width=\linewidth]{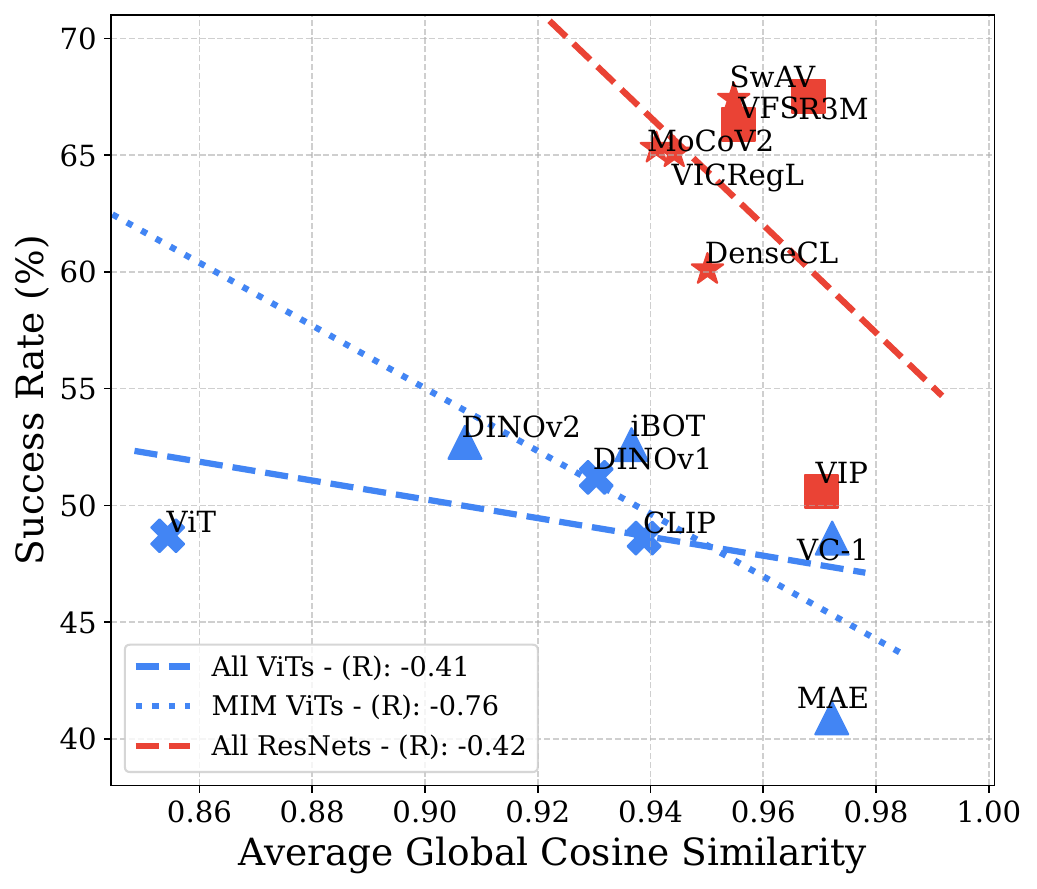}
    \end{subfigure}
    \caption{Correlation plots between per-PVR average policy success rate and temporal entanglement. The left plot concerns short-range temporal entanglement, as described in Section~\ref{ssec:short_range_entanglement} and the right one concerns long-range temporal entanglement, as described in Section~\ref{ssec:long_range_entanglement} (we omit here the non-temporal ResNets sub-group, as no trend emerged).}
    \label{fig:correlations_entanglement} 
    \vspace{-0.99em}  
\end{figure*}

Similarly to the correlation results from section~\ref{ssec:long_range_entanglement}, a trend emerges where long-range temporal entanglement predicts the policy performance, which is visualised in the right plot of Fig.~\ref{fig:correlations_entanglement}. 
This evidence supports our hypothesis that entanglement occurs throughout the trajectory and not only within a short-range of the current timestep. 
This finding is important since most methods, that work either in the latent space (\eg FLARE~\cite{NEURIPS2021_ba3c5fe1}, concatenating prior tokens~\cite{parisi2022unsurprising}, etc.) or in the policy side (\eg LSTMs, Causal Transformers, etc.), aim to disentangle mostly locally, rather than globally. 

\subsection{General Temporal Entanglement}
\label{ssec:general_entanglement}
Each metric introduced in Sections~\ref{ssec:short_range_entanglement},~\ref{ssec:long_range_entanglement} sheds light to part of the problem.
We argue that a latent space overcomes both short and long-range entanglement if it is suitable for accurately perceiving the task-progression during deployment (\ie how close the policy is to solving the task).

For this, we probe the available PVRs with a shallow MLP, training it to predict the task-progression percentage, assuming that $N_T$ corresponds roughly to the completion of the task. 
Then, using the test seeds utilized for evaluating the policies, we measure the \textit{task-progression loss}, which is the mean-squared error between the predicted and ground-truth task-progression percentage.  

\begin{figure*}[t] 
    \centering
    \begin{subfigure}[b]{0.495\textwidth} 
        \centering
        \includegraphics[width=\linewidth]{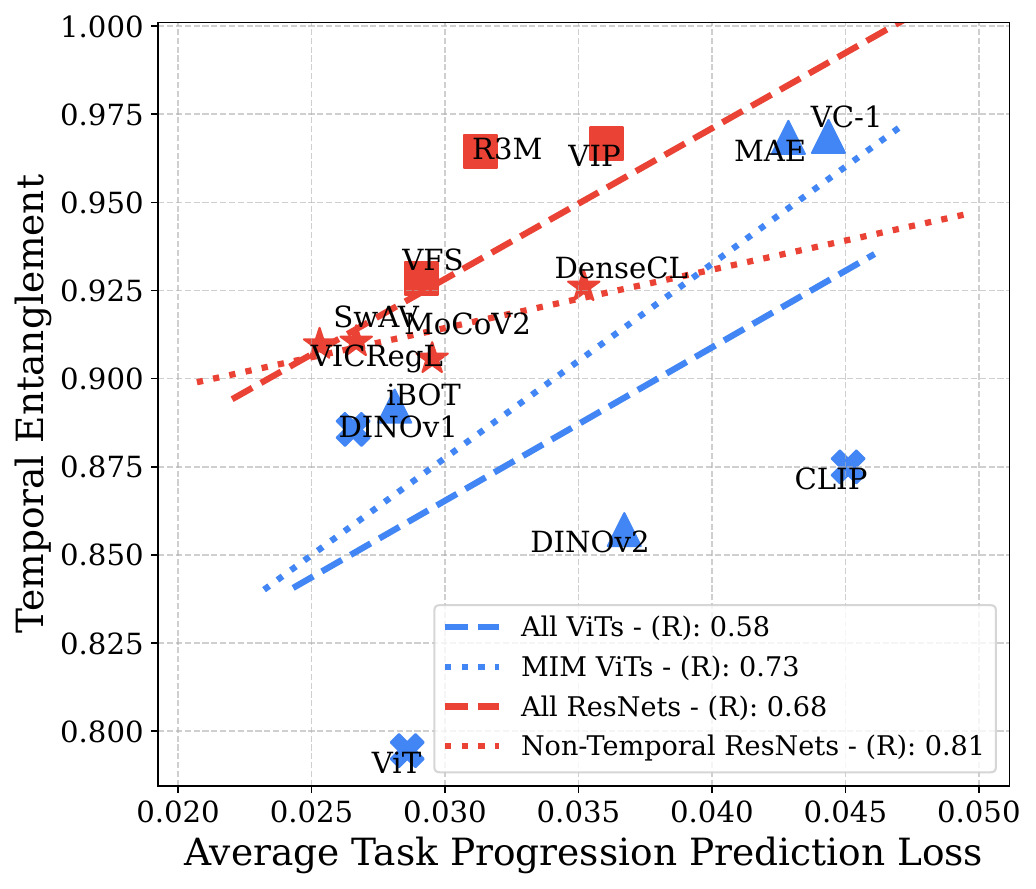}
    \end{subfigure}
    \hfill 
    \begin{subfigure}[b]{0.495\textwidth} 
        \centering
        \includegraphics[width=\linewidth]{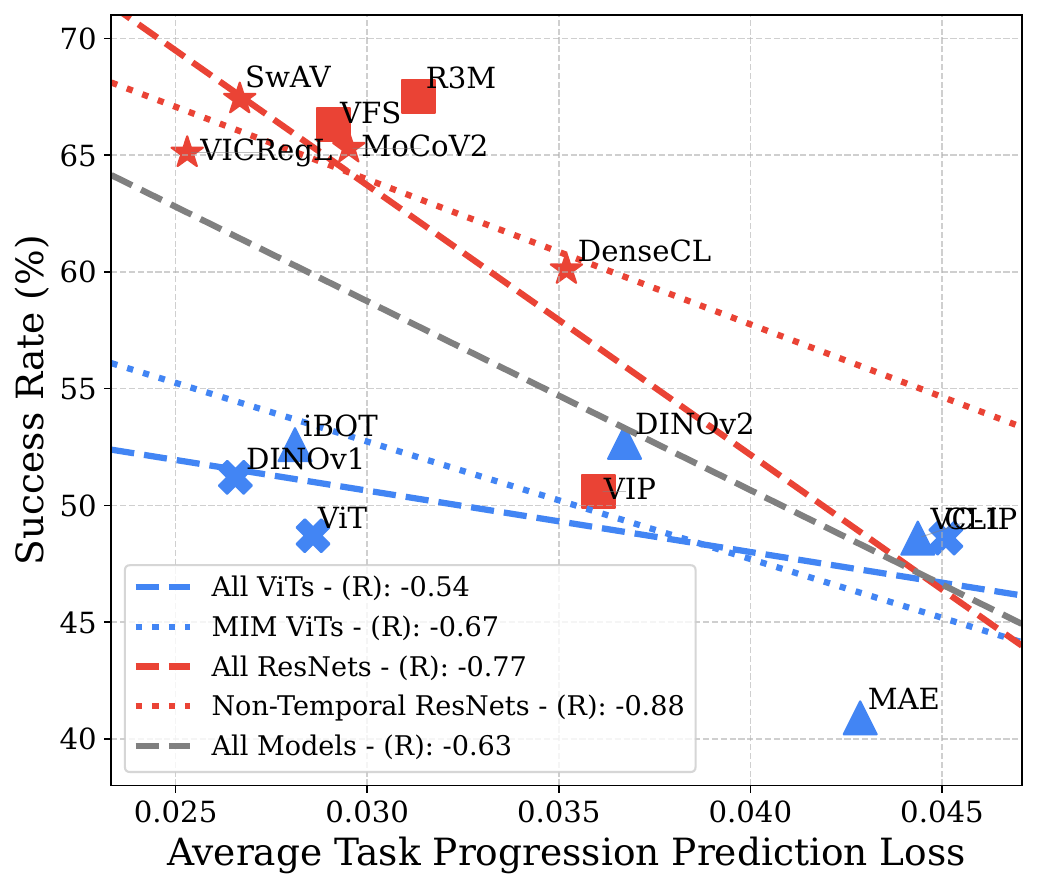}
    \end{subfigure}
    \caption{Correlation plots for the per-PVR task-progression perception. On the left, we provide evidence that good task-progression perception correlates with reduced temporal entanglement. On the right, we showcase that task-progression perception is a good predictor of policy performance (gray line shows trend for all PVRs). }
    \label{fig:correlations_task_progression} 
\vspace{-0.99em}  
\end{figure*}

First, we measure for each PVR the correlation between the task-progression loss and the product of the short and long-range similarity scores, to validate that our proposed probing methods is representative of both types of entanglement metrics. 
Indeed, as visualised in the left plot of Fig.~\ref{fig:correlations_task_progression}, the entanglement scores for the different PVR sub-groups are strongly correlated with the task-progression loss. 
This indicates that we can deploy this tool for measuring the general entanglement (\ie both local and global). 

Second, we measure the correlation between the task-progression loss and the average policy success rate for each PVR.
We discover a strong negative correlation between the two metrics, as visualised in the right plot of  Fig.~\ref{fig:correlations_task_progression}.
We interpret this result as proof that if the PVR latent space is suitable for providing a task-progression signal in the policy network, the success rate is more likely to be high. 

\begin{figure}[th]
    \centering
    \includegraphics[width=1\linewidth]{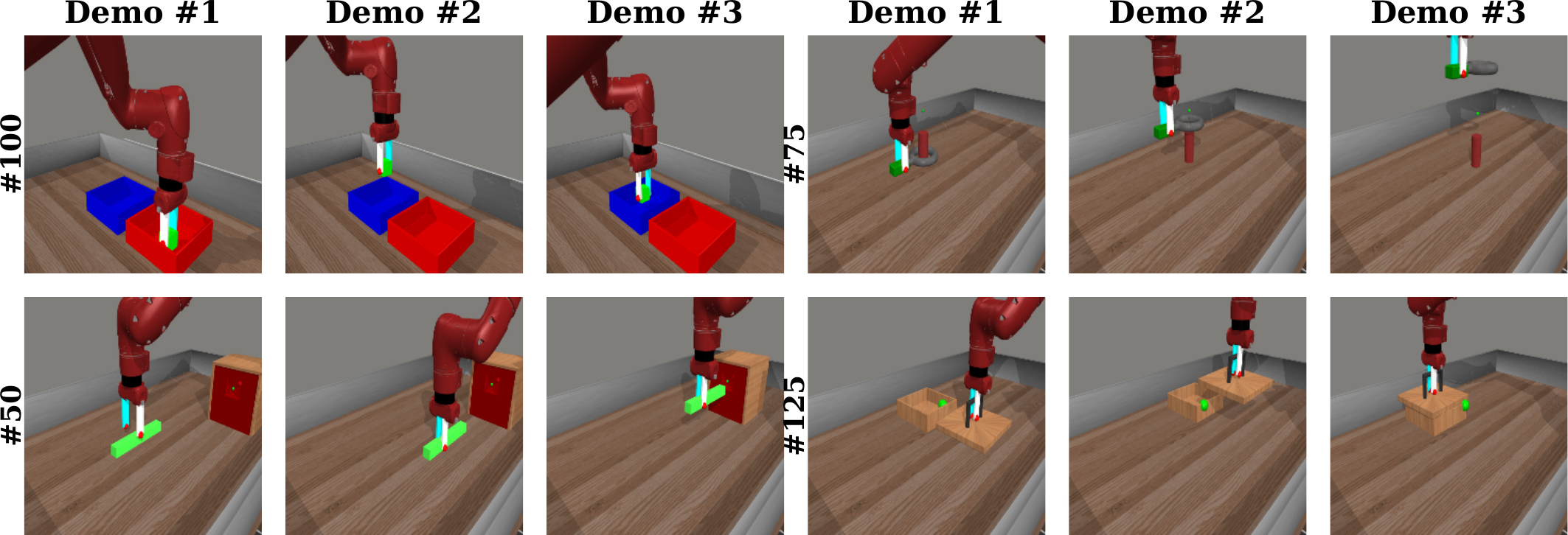}
    \caption{Temporal variability in MetaWorld expert demonstrations: asynchronous task progression is evident in same-time-step frames from separate demonstrations.}
    \label{fig:asynchronous_demos}
\end{figure}

\section{TEMPORAL DISENTANGLEMENT}
\label{sec:experiments}
We leverage our findings from Section~\ref{sec:temporal_entanglement} to better understand how we can effectively disentangle PVR-extracted features in visuomotor policy learning. 
First, in Section~\ref{ssec:te_baseline} we propose a baseline disentangling method, that injects a task-progression signal in the policy model's input.
Then, in Sections~\ref{ssec:feat_aug},~\ref{ssec:causal_transformer} and~\ref{ssec:video_pvrs}, we revisit traditional temporal disentanglement methods, comparing them against our proposed baseline.  

\subsection{Temporal Disentangling Baseline}
\label{ssec:te_baseline}
Our experiments from Section~\ref{ssec:general_entanglement} indicated that the more suitable a PVR's latent space is for task-progression perception, the more likely it is to score high policy success rate. 
Motivated by this observation, we propose a simple yet very effective approach to augment each observation with a temporal component, by encoding the timestep index $n$ of each frame as a high-dimensional vector, using: 

\begin{align}
\label{eq:t_enc}
\gamma(n) &= \Big( 
\sin\left( \frac{2^0 \pi n}{s^0} \right), \cos\left( \frac{2^0 \pi n}{s^0} \right), \dots, \nonumber\\
&\quad \sin\left( \frac{2^{N_T-1} \pi n}{s^{N_T-1}} \right), \cos\left( \frac{2^{N_T-1} \pi n}{s^{N_T-1}} \right) 
\Big)
\end{align}


\noindent and concatenating to the policy input (see Fig.~\ref{fig:policy}). 
This augmentation can temporally disentangle similar $(p_t, o_t)$ pairs, introducing a task progression signal into the robot state, which we argue can enhance policy performance.
We empirically set the dimensionality of the temporal embeddings to be $64$ and the scale parameter $100$.

Using alternating sine and cosine functions at exponentially increasing frequencies \( 2^k \), the lower-frequency terms capture coarse temporal trends, while the higher-frequency terms provide finer temporal resolution, enabling the policy to distinguish between temporally similar states in both short and long range. 
Note that traditionally such embeddings encode the relative position in a transformer's input~\cite{su2024roformer}, whereas here we encode the position of the embedding in the rollout.

While our approach currently has limitations that preclude it as a real-world solution (\eg policy cold-starts may be difficult), we believe it serves as a strong baseline for future disentanglement techniques.

\subsection{Feature augmentation-based disentanglement}
\label{ssec:feat_aug}
We first investigate the approach of temporally augmenting the PVR-features, by concatenating the three most recent past observations and their latent differences (\ie FLARE~\cite{NEURIPS2021_ba3c5fe1}).
Even though this decreases the PVR short-range temporal entanglement from 0.9686 to 0.8904, the average success rate increases by only roughly 2\%, from 56.11\% to 58.34\% (detailed results in Fig.~\ref{fig:results}).
On the other hand, our baseline (PVR+TE) demonstrates a decrease in short-range temporal entanglement to only 0.9334, but the increase of the average PVR success rate is more than 7\% (63.35\%).
This can be explained if we observe the long-range similarity of features.
FLARE reduces it from 0.9426 to 0.9377, while TE scores 0.1520. 
Note, that as visualised in Fig.~\ref{fig:asynchronous_demos}, MetaWorld expert demos exhibit temporal variability, and thus the increased performance should not be attributed to overfitting to \textit{(time-index, action)} pairs, but rather to the injected progression signal. 

These results further underscore that extracting a temporal signal from only nearby observations might not be enough, hinting that a more intricate solution is required that provides a sense of task-progression. 
An intriguing observation is that PVRs pre-trained with a temporal component in their objective function also benefit considerably from TE. 
For instance, R3M employs time-contrastive learning~\cite{sermanet2017unsupervised} to enforce similarity between representations of temporally adjacent frames-experience substantial gains from TE. 
In addition, VIP achieves an average performance boost of approximately $15$\%, making it one of the most positively impacted models. 
This finding suggests a potential reconsideration of how temporal perception is integrated into features designed for robot learning. 
It raises the possibility that existing approaches may not fully exploit the temporal structure necessary for optimal performance.
Overall, while VC-1 and iBOT achieve slightly higher average scores with FLARE, all other PVRs benefit significantly from temporal augmentation, even when compared to FLARE-augmented results. 
Similarly, apart from the ``Pick and Place'' and ``Shelf Place'' tasks, TE significantly enhances the average task performance.
Conducting a statistical analysis we confirmed that our TE baseline's gains over FLARE and no augmentation are significant.
This is supported by both the Wilcoxon test ($p < 10^{-30}$) and paired t-test ($p < 10^{-26}$). 
Additionally, TE also outperforms FLARE, with statistically significant differences observed in both the Wilcoxon test ($p \approx 4.38 \times 10^{-5}$) and paired t-test ($p \approx 1.47 \times 10^{-4}$).

\begin{figure}[ht]
    \centering
    \begin{tabular}{c}
        \includegraphics[width=0.95\linewidth]{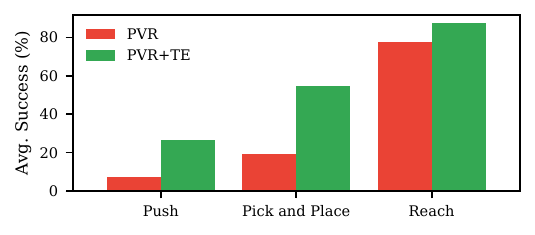} \\
        \includegraphics[width=0.95\linewidth]{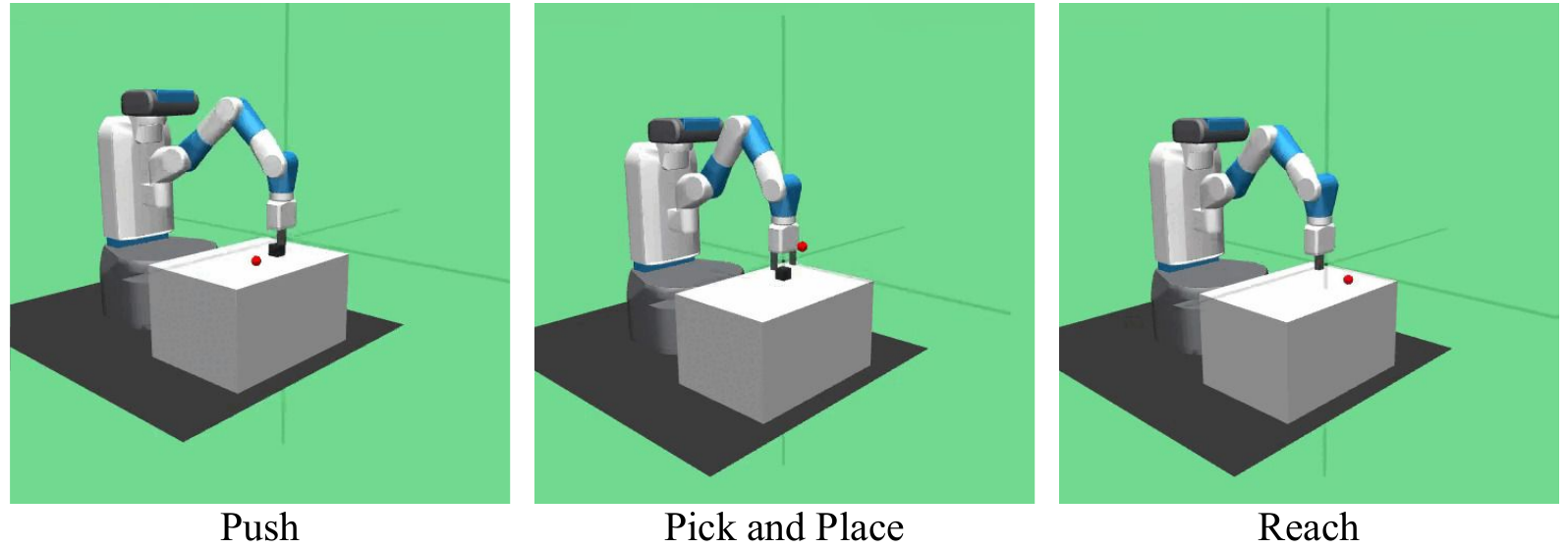}
    \end{tabular}
    \caption{Validation results of the TE baseline on three additional tasks from the OpenAI Robotics Suite.}
    \label{fig:openai}
    \vspace{-0.99em} 
\end{figure}

Finally, to make sure that our findings are general and not simulation-specific, we validate that PVR+TE improved the policy performance in tasks from a different simulation environment.
We use the \texttt{Push}, \texttt{Pick and Place} and \texttt{Reach} tasks from the OpenAI Robotics Suite (visualised in Fig.~\ref{fig:openai}) and 50 expert demonstrations per task from~\cite{haldar2022watch}.
We present the average per-task success rates in Fig.~\ref{fig:openai}, where TE more than doubles performance on the \texttt{Push} and \texttt{Pick and Place} tasks, and still delivers a notable, though smaller, improvement on \texttt{Reach}, likely due to the task’s lower difficulty.

\subsection{Policy-based disentanglement}
\label{ssec:causal_transformer}

We revisit the idea of disentangling features by relying on a policy architecture that captures temporal relationships.
We deploy a 1-layer Causal Transformer (CT)~\cite{zhou2025dinowmworldmodelspretrained, fu2024icrt,huang2025otter}, which encodes temporal relationships both in the input and in the output, by utilising context and action chunking of length $12$, and train it with the features of the strong-performing R3M for $0.8$M steps on four MetaWorld tasks.
We choose to test our CT in a multi-task setting to further validate TE's ability to temporally disentangle similar features, rather than encode absolute timesteps.

Indeed, Table~\ref{tab:causal_t} reveals that augmenting the input feature space with a task-progression signal benefits greatly the policy in successfully completing a task, leading to an average task success rate boost of +20\%. 
 
\begin{table}[t]
    \centering
    \caption{Multi-task success rate (Peg Insert, Bin Picking, Disassemble, Coffee Pull) of a Causal Transformer trained with and without TE.}
    \begin{tabular}{lccccc}
        \toprule
        & \textbf{Task 1} & \textbf{Task 2} & \textbf{Task 3} & \textbf{Task 4} & \textbf{Average} \\
        \midrule
        \textbf{CT}       & 42\%  & 80\%  & 54\%  & 96\%  & 68.0\%  \\
        \textbf{CT+TE}  & 62\%  & 90\%  & 93\%  & 100\% & 86.3\%  \\
        \bottomrule
    \end{tabular}

    \vskip -0.1in
    \label{tab:causal_t}
    
\end{table}

\subsection{PVR-based disentanglement}
\label{ssec:video_pvrs}
Finally, we challenge the starting point of this work, which was to rely on PVRs meant for static image processing, even if some of them were trained with frames extracted from video datasets. 
We explore the idea that potentially PVRs inherently struggle to deal with temporal entanglement due the nature of their training data and explore the idea of relying on Video-PVRs (\ie pre-trained models that were meant for video applications, and thus potentially encode implicitly temporal structures).
For this purpose, we deploy three powerful video encoders (TimeSformer~\cite{timesformer}, ViViT~\cite{ViViT} and VideoMAE~\cite{tong2022videomae}) that have all been trained with the Kinetics-400 dataset~\cite{kay2017kineticshumanactionvideo} and utilise the ViT-B/16 backbone. 

In these experiments we preserve our methodology, apart from the way the frame input stream is processed. 
In the case of video encoders, for both training and evaluating, a frame buffer is created of length $N_f$, which has the most recent frame, followed by the $N_f-1$ previous ones. 
Until the number of available frames become equal to $N_f$, the buffer is padded, by repeating the oldest frame. 
We set $N_f$ to match the frame history length of each encoder.

\begin{table}[b]
    \centering
    \caption{Comparison of inference time and input frames across three video encoders versus the vanilla ViT-B/16, which processes a single frame and is the base for all our ViT-based PVRs except DINOv2 (patch size 14).}
    \begin{tabular}{lcccc}
        \toprule
        & \textbf{ViT-B/16} & \textbf{TimeSformer} & \textbf{VideoMAE} & \textbf{ViViT} \\
        \midrule
        $\boldsymbol{N_f}$ & 1 & 8 & 16 & 32 \\ 
        $\boldsymbol{T_p}$ & $\approx 0.025\text{s}$ & $\approx 0.145\text{s}$ & $\approx 0.265\text{s}$ & $\approx 0.550\text{s}$ \\ 
        \midrule
        \textbf{VPVR}         & -- & 56.9\% & 45.5\% & 18.8\% \\
        \textbf{VPVR+TE}    & -- & 62.4\% & 44.8\% & 24.9\% \\
        \bottomrule
    \end{tabular}
    \label{tab:video_enc}
\end{table}


Table~\ref{tab:video_enc} summarises three important aspects of pre-trained video encoders. 
First, Video-PVRs seem to struggle to outperform even the average PVR performance.
Second, we measure the average inference time of each encoder and compare it against the time it takes to process a single frame for the same backbone (\ie ViT-B/16), which is the one utilised by other PVRs in our experiments. 
It is not a surprise that the larger $N_f$ is, the slower inference gets~\footnote{Note that all preprocessing modules and model inference times were measured using code from~\texttt{\href{https://huggingface.co/docs/transformers/main/index}{huggingface.co/docs/transformers}} and tested on a NVIDIA GeForce RTX 4090 GPU with 24GB VRAM, using batches of size 25.}. 
Finally, an interesting find concerns the average success rate itself, which seems to be negatively correlated with the number of frames in the buffer. 
This counter-intuitive result aligns with the findings of~\cite{chi2023diffusionpolicy}, regarding the length of the observation horizon, where the performance would decline as the length increased.   

\section{CONCLUSIONS}
Our findings suggest that temporal entanglement and a lack of task-progression perception are imperative limitations that accompany the deployment of PVRs in visuomotor policy learning. 
Our proposed baseline provides evidence that current techniques that aim to enhance the policy input with a temporal component fall short.
At the same time, PVRs trained with objective functions and dataset that contain a temporal component also seem to show room for improvement. 
We believe that our findings can help pave the way towards a temporally aware PVR that will mitigate the investigated issues. 









\bibliographystyle{plain} 
\bibliography{references}
\end{document}